\documentclass{article}

\usepackage[final]{nips_2018}

\usepackage{graphicx}
\usepackage[utf8]{inputenc} 
\usepackage[T1]{fontenc}    
\usepackage{hyperref}       
\usepackage{url}            
\usepackage{booktabs}       
\usepackage{amsfonts}       
\usepackage{nicefrac}       
\usepackage{microtype}      
\usepackage{times}
\usepackage{latexsym}
\usepackage{url}
\usepackage{bm}
\usepackage{amsmath, amssymb}
\usepackage{algorithm, algorithmic}
\usepackage{todonotes}
\usepackage{titlesec}
\usepackage{pgfplots}
\usepackage{float}
\usepackage{authblk}

\title{Cross-Lingual Approaches to Reference Resolution in Dialogue Systems}
\author[1]{\bf Amr Sharaf}
\author[2]{\bf Arpit Gupta}
\author[2]{\bf Hancheng Ge}
\author[2]{\bf Chetan Naik}
\author[2]{\bf Lambert Mathias}
\affil[1]{University of Maryland}
\affil[2]{Amazon Alexa AI}
\affil[1]{amr@cs.umd.edu}
\affil[2 ]{\{arpgup,ghanche,chetnaik,mathiasl\}@amazon.com}
\date{}

\begin{document}
\maketitle
\begin{abstract}
In the slot-filling paradigm,  where a user can refer back to slots in the context during the conversation, the goal of the contextual understanding system is to resolve the referring expressions to the appropriate slots in the context. In this paper, we build on~\citep{Naik2018ContextualSC}, which provides a scalable multi-domain framework for resolving references. However, scaling this approach across languages is not a trivial task, due to the large demand on acquisition of annotated data in the target language. Our main focus is on cross-lingual methods for reference resolution as a way to alleviate the need for annotated data in the target language. In the cross-lingual setup, we assume there is access to annotated resources as well as a well trained model in the source language and little to no annotated data in the target language. In this paper, we explore three different approaches for cross-lingual transfer \textemdash~\ delexicalization as data augmentation, multilingual embeddings and machine translation. We compare these approaches both on a low resource setting as well as a large resource setting. Our experiments show that multilingual embeddings and delexicalization via data augmentation have a significant impact in the low resource setting, but the gains diminish as the amount of available data in the target language increases. Furthermore, when combined with machine translation we can get performance very close to actual live data in the target language, with only 25\% of the data projected into the target language.

 \end{abstract}

\section{Introduction}
\begin{figure}[h]
  \begin{center}
    \includegraphics[width=7.5cm]{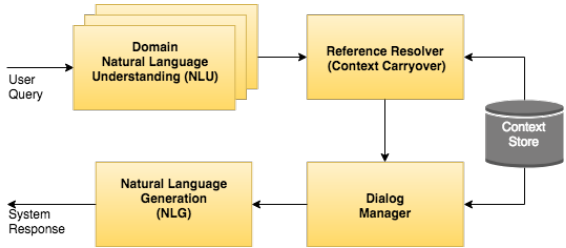}
    \caption{Spoken Dialogue Architecture: A pipelined approach for spoken language understanding. The context carryover system is responsible for resolving references in a conversation and is an input to the dialogue manager.}
    \label{fig:slu_arch}
  \end{center}
\end{figure}

Most commercial spoken language systems consist of multiple components~\citep{tur2011spoken}. Figure~\ref{fig:slu_arch} describes this architecture -  a user query is presented as text (this could be the output of a speech recognizer in case of spoken language input) to a collection of domain-specific natural language understanding (NLU) systems, which produces the most likely semantic interpretation - typically represented as intents and slots~\citep{wang2011semantic}. The resulting semantic interpretation is then passed to a reference resolver, in order to resolve referring expressions (including anaphora) to their antecedent slots in the conversation, which is then sent to the dialogue manager whose main responsibility is to determine the next action. This action is the input to the natural language generation component that is responsible for generating the system response back to the user. 

In this paper, we focus on the reference resolution task. Resolving anaphora and referring expressions is an important sub-component and is essential for maintaining the state of the conversation across turns~\citep{celikyilmaz2014resolving}. The key here is to leverage the dialogue context effectively~\citep{Bhargava2013EasyCI, Xu2014ContextualDC} for improving spoken language understanding accuracy. However, in commercial systems like Siri, Google Home and Alexa, the NLU component is a highly federated and diverse collection of services spanning rules or diverse statistical models. Typical end-to-end approaches~\cite{bapna2017sequential} which require back-propagation through the NLU sub-systems are not feasible in such a setting, effectively the NLU systems could be considered immutable.  This issue is addressed in \citep{Naik2018ContextualSC}, which describes a scalable multi-domain architecture for solving the reference resolution problem called {\it context carryover}, and does not require coupling with the multiple NLU sub-systems. However, the focus of that work is mainly on the monolingual setting where annotated resources are readily available. In this paper, our main contribution is to show how {\it context carryover} can be extended to new languages effectively using cross-lingual transfer. 

While a simple naive approach would be to collect annotated resources in the target language, this is often expensive and time consuming. In the cross-lingual setting, we assume we have access to annotated data in the source language (en\_US) on which we can train the context carryover model. For the target language (de\_DE), we assume we have little to no annotated data available, which is typically the case in a low-resource setting. In this paper, we empirically investigate three approaches for cross-lingual transfer, multilingual embeddings, translation projection and delexicalization.  We show that translation is an effective strategy to bootstrap a model in the target domain.  Furthermore, we demonstrate that data augmentation strategies that abstract from the lexical representation can significantly boost the performance of the models, especially in the low resource setting.

This paper is structured as follows. In Section~\ref{ssec:task}, we give a formal description of the carryover task. In Section~\ref{ssec:crosslingual}, we outline multiple approaches to 
cross-lingual transfer. In Section~\ref{sec:experiments}, we present empirical results comparing and analyzing the effectiveness of multiple strategies for cross-lingual training of these 
models. Finally, we present our conclusions and a few possible areas of further research.

\section{Approach}
\label{sec:approach}

\subsection{Task Definition}
\label{ssec:task}
We first motivate the context carryover task with the example shown in Table~\ref{tbl:cca}. At turn {\bf U3}, the context carryover system resolves the implicit reference by determining that the contextual slots ({\it Place=Exploratorium}) and ({\it City=san francisco}) are most relevant to the user utterance  "{\it what's the address}", but ({\it PlaceType=museum}) is no longer relevant. During training the model learns the carryover action by leveraging the similarities between the candidate slot and the dialogue context. At test time, the context carryover system is activated only on a user turn for which dialog context is available, and the decision for each slot is made independently. 

\begin{table*}[ht]
\begin{center}
\scalebox{1.0}{
\begin{tabular}{|c|c|c|c|}\hline                                                                                                                                                                                                                         
\multicolumn{1}{|c|}{\bf Domain} & \multicolumn{1}{|c|}{\bf Turns} & \multicolumn{1}{|c|}{\bf Current Turn Slots} & \multicolumn{1}{|c|}{\bf Carryover Slot} \\
\hline\hline
Local & {\bf U1}: {\text find a museum in san francisco} & {\text PlaceType=museum} & - \\
& & {\text City=san francisco} &  \\\hline
Local & {\bf V1}: Found exploratorium it & {\text Place=Exploratorium}  \\
& is 10 miles away & {\text Distance=10 miles} &\\\hline
Local & {\bf U2}: what's the address & &{\text Place=Exploratorium}  \\
& \textit{\textbf{<Implicit Reference>}} & &{\text City=san francisco} \\\hline
Local & {\bf V2}: {\text located on Embarcadero st..} & {\text Location=Pier 15,} & \\
& & Embarcadero St & \\\hline
Calling & {\bf U3}: {\text call them} & &{\text Contact=Exploratorium} \\
&  \textit{\textbf{<Explicit Reference>}} & & \\\hline
\end{tabular}
}
\end{center}
\caption{Context Carryover resolves explicit and implicit references in the dialogue - at each turn the most relevant slot from the context is 'carried over' to the current turn. The 
conversation from U2 to U3 involves a domain change - Local to Calling - and a schema change i.e the slot  Place from the Local domain needs to be carried over and transformed to  Contact in the Calling domain.}
\label{tbl:cca}
\end{table*}

We can now formally define the context carryover task. We define a dialogue turn at time $t$ as the tuple $\{a_t, \bm{S}_t, \bm{w}_t\}$, where $\bm{w_t} \in \mathcal{W}$ is  a sequence 
of words $\{w_{it}\}_{i=1}^{N_t}$; $a_t \in \mathcal{A}$ is the dialogue act; and $\bm{S}_t$ is a set of slots, where each slot $s$ is a key value pair $s=\{k,v\}$, with $k\in \mathcal{K}$ being 
the slot name (or slot key), and $v\in \mathcal{V}$ being the slot value. $\bm{u}_t=\{a_t^u, \bm{S}_t^u, \bm{w}_t^u\}$ represents a user-initiated turn and $\bm{v}_t=\{a_t^v, \bm{S}_t^v, 
\bm{w}_t^v\}$ represents a system initiated turn. Given a sequence of $D$ user turns $\{\bm{u}_{t-D}, \dots, \bm{u}_{t-2}, \bm{u}_{t-1}\}$; and their associated system turns $\{\bm{v}
_{t-D}, \dots, \bm{v}_{t-2}, \bm{v}_{t-1}\}$\footnote{For simplicity we assume a turn taking model - a user turn and system turn alternate.}; and the current user turn $\bm{u}_t$, we  
construct a candidate set of slots from the context 
\begin{equation}
C(\bm{S})=\bigcup\limits_{i \in {u,v}, j=t-D}^{t-1} \bm{S}^i_j
\end{equation}
 
\begin{figure*}[ht]
  \begin{center}
    \includegraphics[scale=0.35]{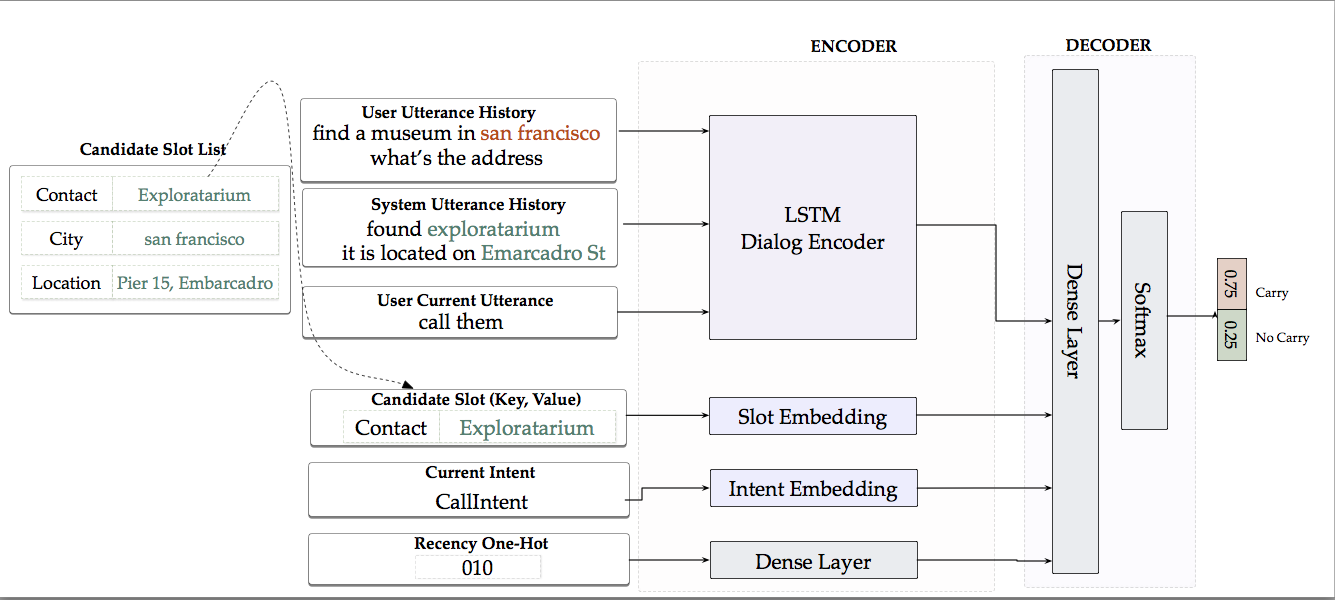}
    \caption{Context Carryover Architecture: Resolving slot references across a collection of diverse NLU sub-systems.}
    \label{fig:cc_arch}
  \end{center}
\end{figure*}

For a candidate slot $s \in C(\bm{S})$, for the dialogue turn at time $t$,  we formulate context carryover as a binary classification task
\begin{equation}
P(+1|s, \bm{u}_t; \bm{u}_{t-D}^{t-1}, \bm{v}_{t-D}^{t-1}; d(s)) > \tau
\end{equation}

where, $\tau$ is a tunable decision threshold. $d(s) \in [0,D]$ is an integer value describing the offset of the slot from the current turn $\bm{u}_t$. We use a neural network encoder-decoder formulation to model this task as shown in Figure~\ref{fig:cc_arch}. For a detailed description of the model, the reader is referred to~\citep{Naik2018ContextualSC}.

\subsection{Cross-Lingual Model Transfer}
\label{ssec:crosslingual}
In order to alleviate the cost of obtaining annotated resources in a new language, we focus on cross-lingual techniques. In the cross-lingual setting, a model is trained on a source language with sufficiently large annotated data, and the learnt information is then transferred to a target language with low resources or even zero training data~\citep{pan2010survey,kozhevnikov2016cross}. In this work, we compare and contrast 3 approaches to cross-lingual transfer.

\subsubsection{Multilingual word embeddings}
We initialize the models with multilingual word embeddings where the words in various languages have been projected into a common language space. Here, we leverage the fasttext embeddings as described in \citep{lample2017unsupervised}. The hypothesis is that because the words across languages that share the same semantics are in the same vector space, training on the source language should result in effective transfer to the target language without having to rely on a lot of target language data.

\subsubsection{Translation projection}
Similar to the work done in \citep{lefevre2010cross}, we use a statistical machine translation system to translate the source language into the target language, and then we train the model in the translated target language. Although the training data is likely to include errors due to translation inaccuracy, this  method  can  potentially  predict  the  correct semantics by learning from consistent error patterns.

\subsubsection{Delexicalization}
\label{sssec:delex}
We follow the approach outlined in \citep{henderson2014robust} where we replace the input streams to the model - the context turns, the current turn and the slots - with generic symbols representing the slot key and the intent patterns associated with those turns, as shown in~\ref{tbl:delex}. We augment the {\it original} training data with this delexicalized data. This allows the model to generalize to unseen lexical tokens, by focusing the model via the attention mechanism on the semantic representation.

\begin{table*}[ht]
\begin{center}
\scalebox{0.75}{
\begin{tabular}{|c|c|c|c|}\hline                                                                                                                                                                                                                         
Turns & Intent & Slots & Delexicalized Turn\\\hline\hline
{\bf U1}: {\text find a museum in san francisco} & Local.SearchPlaceIntent & {\text PlaceType=museum} & {\text Local.SearchPlaceIntent find a PlaceType} \\
& & {\text City=san francisco} & {\text in City} \\\hline
{\bf V1}: Found exploratorium it & Local.InformAction & {\text Place=Exploratorium}  & {\text Local.InformAction found Place}\\
 is 10 miles away & & {\text Distance=10 miles}  & {\text it is Distance away.}\\\hline
\end{tabular}
}
\end{center}
\caption{The user and system utterance tokens are delexicalized by replacing the tokens with the slot keys and inserting the intent label at the beginning.}
\label{tbl:delex}
\end{table*}

\section{Experiments}
\label{sec:experiments}
\subsection{Data Setup}
For the experiments, we present our results on a sampled internal benchmark dataset for voice based applications that spans six domains \textemdash~\ weather, music, video, local business search, movie showtimes and general question answering. Table~\ref{tbl:data} summarizes the statistics in the training, development and test sets across different domains. We focus on en\_US and de\_DE as the two languages for all our multilingual experiments. Both datasets have very similar distribution of average number of user turns. 
\begin{table*}[ht]
\begin{center}
\begin{tabular}{|c||c|c|c|c|c|c|} \hline
& \multicolumn{3}{|c|}{en\_US} &  \multicolumn{3}{|c|}{de\_DE}\\\hline
& Train & Dev & Test & Train & Dev & Test \\\hline
 Number of Sessions & 18k & 15k  & 15k & 17k & 13k & 12k\\
Average turns per session & 2.2 & 2.14 & 2.18 & 2.20 & 2.0 & 2.17 \\\hline
\end{tabular}
\end{center}
\caption{Context Carryover Data Setup}
\label{tbl:data}
\end{table*}

\subsubsection{Training Setup and Evaluation Metrics}
\label{ssec:setup}

For the model, we initialize the word embeddings using fasttext embedings~\citep{lample2017unsupervised}. The model is trained using mini-batch SGD with Adam optimizer~\citep{Kingma2014AdamAM} with standard parameters to minimize the class weighted cross-entropy loss. In our experiments, we use 128 dimensions for the LSTM hidden states and 256 dimensions for the hidden state in the decoder. Similar to~\citep{Wiseman2015LearningAA}, we pre-train an LSTM  encoder using live data in each of the languages and use this model to initialize the parameters of the LSTM-based encoders. All model setups are trained for 40 epochs. For evaluation on test set we pick the best model based on performance on dev set.  For evaluation, we only select those slots as the final hypothesis, whose $\tau > 0.5$ and occur in the context of the conversation. For each utterance, independent carryover decisions are taken for each candidate slot. We use standard definitions of precision, recall and F1 by comparing the reference slots with the model hypothesis slots. If an entity type is repeated in the current turn then we do not carry this from dialogue history.

\subsection{Establishing the monolingual baselines}
For comparison, we introduce a simple naive baseline that carries over the most recent slots in the dialogue session, that demonstrates the complexity of the task - simply carrying over the slot from the context results in very poor performance in the context carryover task. Table~\ref{tbl:results_en} establishes the monolingual baseline for en\_US and de\_DE based on available annotated data described in Table~\ref{tbl:data}, and using the neural network architecture described earlier.

\begin{table*}[ht]
\centering
\begin{tabular}{|c||c|c|c|c|c|c|} \hline
& \multicolumn{3}{|c|}{en\_US} &  \multicolumn{3}{|c|}{de\_DE}\\\hline
Method & Precision & Recall & F1 & Precision & Recall & F1\\\hline
Baseline & 37.77 & 93.75 & 53.85  & 30.57 & 85.66 & 45.06\\
Encoder-Decoder w/ attention & 90.7 & 93.1 & 91.9 & 77.6 & 69.9 & 73.07\\\hline
\end{tabular}
\caption{Monolingual Baseline Experiments for en\_US and de\_DE. The train and test sets are on fully annotated live datasets. For de\_DE this represents the best performance when annotated data is available in the target domain.}
\label{tbl:results_en}
\end{table*}



\subsection{Cross-lingual Transfer Experiment Results}
Training a contextual slot carryover model across each locale independently is challenging, primarily due to the cost of acquiring labeled conversational data. Instead, we focus on methods for leveraging cross-lingual information. For all the experiments, we tune the model on the de\_DE development set, and evaluate the results on the de\_DE test. This dev and test data composition is the one described in Table~\ref{tbl:data}.

\subsubsection{Impact of multilingual embeddings}
Here we compare the monolingually trained fasttext embeddings~\citep{bojanowski2016enriching} to the multilingually aligned embeddings~\citep{conneau2017word}. In order to take advantage of the multilingual embeddings we train the model jointly on en\_US data and en\_US data translated to de\_DE data, and evaluate the models on true de\_DE test data. We find that multilingual embeddings help in the low resource setting, but as the available training data increases, the gains diminish~\footnote{In our setup, we also found that multilingual embeddings only help when doing multilingual training i.e jointly training across en\_US and de\_DE}.
\begin{table*}[ht]
\centering
\begin{tabular}{|c|c|c|c|c|} \hline
\% translated en\_US->de\_DE & Embedding & Precision & Recall & F1 \\\hline
1\% & Monolingual & 63.9   &  60.0  & 61.6\\
 & Multilingual & 62.5  &  64.2 & {\bf 63.3}  \\\hline
25\% & Monolingual & 72.5    & 59.6  & 63.1 \\
& Multilingual & 73.5  &  62.8 & {\bf 66.4}\\\hline
100\% & Monolingual & 71.2   & 63.1  & 66.0 \\
& Multilingual & 72.9  &  62.5 & 65.9 \\\hline
\end{tabular}
\caption{Impact of multilingual embeddings on test de\_DE data: Multilingual embeddings provide diminishing gains as the amount of de\_DE training data increases.}
\label{tbl:results_translation}
\end{table*}

\subsubsection{Impact of Translation Projection}
Here we use an in-house en\_US to de\_DE translation system. We translate all en\_US data described in Table ~\ref{tbl:data}. In order to estimate the translation quality, we perform back translation from de\_DE to en\_US; the resulting translation when compared to the reference input en\_US utterances gives us a BLEU~\citep{papineni2002bleu} score of 29.36.

\begin{table*}[ht]
\centering
\begin{tabular}{|c|c|c|c|} \hline
\% translated en\_US->de\_DE & Precision & Recall & F1 \\\hline
1\% & 50.76 & 51.5 & 50.44 \\\hline
25\% & 70.5 & 60.4 & 63.6 \\\hline
100\% & 72.1 & 63.8 & 66.8\\\hline 
\end{tabular}
\caption{Impact of translation projection on live de\_DE test data for varying amounts of translated en\_US->de\_DE data. Compared to the monolingual de\_DE baseline in Table~\ref{tbl:results_en}, translation projection is within 10\% relative F1, indicating that the noise in translation results in poorer performance.}
\label{tbl:results_translation}
\end{table*}

From Table~\ref{tbl:results_translation}, we can see that as we increase the amount of translated data, the performance of the models on the target language increases significantly.  While the performance on translation data is still within $10\%$ relative F1 compared to a model trained on live de\_DE data (as shown in Table~\ref{tbl:results_en}), this shows that  translation projection provides a viable alternative to bootstrap context carryover models in a new language. 

\subsubsection{Impact of Delexicalization}
In many cases, if we have access to the source language data, we can leverage it in addition to training on the synthesized translated data in the target language. In this paper, we leverage the available source language en\_US data in two ways. We first consider the impact of delexicalization, described in Section~\ref{sssec:delex}. In this setting, we simply augment the target language de\_DE data with delexicalized data from en\_US. Furthermore, we also study the impact of initializing the model from trained en\_US parameters (we call this source language initialization) vs training from scratch. Table~\ref{tbl:results_delex}, describes in detail the various experiments. We see that for the low resource setting, both delexicalization and source language initialization of the models have a huge impact on the performance on the target language. We get a $26\%$ relative F1 improvement by leveraging these techniques. However, the gains diminish as we increase the amount of target language data. While delexicalization still gives a $4-9\%$ lift in relative F1 scores, we no longer get any gains from initializing the model from en\_US data. Also, we see that we can get most of the gains from only translating $25\%$ of the source language data, which indicates that we can bootstrap a new language with far fewer resources in the target language.

\begin{table*}[ht]
\centering
\small
\begin{tabular}{|c|c|c|c|c|c|} \hline
\% translated en\_US->de\_DE & Delexicalization & Source Language initialization & Precision & Recall & F1 \\\hline
1\% & No & No & 50.76 & 51.5 & 50.44 \\
& Yes & No & 63.7 & 57.0 & 59.0 \\
& Yes & Yes & 64.2 & 62.9 & {\bf 63.6} \\\hline
25\% & No & No & 70.5 & 60.4 & 63.6 \\
& Yes & No & 68.7 & 70.0 & {\bf 69.3} \\
& Yes & Yes & 68.9 & 67.8 & 68.3 \\\hline
100\% & No & No & 72.1 & 63.8 & 66.8\\
& Yes & No & 68.2 & 71.8 & {\bf 69.8}\\
& Yes & Yes & 71.5 & 66.7 & 68.8\\\hline
\end{tabular}
\caption{Impact of translation projection on live de\_DE test data for varying amounts of translated en\_US->de\_DE training data. With only 25\% of the translated data, combining with delexicalization we get significant improvements, within 4\% F1 of the monolingual de\_DE system. Source language initialization does not help except in the low resource setting when the target language data is in the 1\% range.}
\label{tbl:results_delex}
\end{table*}

\section{Related Work}
\label{sec:related}
Most of the approaches for contextual understanding and dialogue have focused on en\_US as the language, due to a wide variety of available resources~\citep{bapna2017sequential, 
chen2016end,Henderson2013DeepNN}. However, there is very little focus on multilingual contextual understanding. Recently, DTSC5 introduced a challenge where data was provided 
for training and developing in en\_US, and the systems were evaluated on Chinese~\citep{williams2013dialog}.~\citep{hori2016dialog} describes a combined en\_US and rule based 
system to solve the task. However, this approach relies more on language agnostic rules, and does not really exploit any other data source in the target language. 

Cross-lingual model transfer consists of modifying a source language model to make it directly applicable to a new language. This usually involves constructing a shared feature 
representation across the two languages. \citep{mcdonald2011multi}   successfully   apply  this  idea  to  the  transfer  of  dependency parsers,   using   part-of-speech   tags   as   the 
shared representation of words.   A later extension  of  \citep{tackstrom2012cross}  enriches this  representation  with  cross-lingual  word clusters,  considerably improving the performance. \citep{kozhevnikov2016cross}  used  cross-lingual   model   transfer   to   learn   a   model for   Semantic   Role   Labeling   (SRL).   The model  combines  both  syntactic  
representations shared across different languages (like universal part-of-speech tags) as well as semantic  shared  representations  using  cross-lingual  word  clusters  and  cross-
lingual  distributed word representations. Closely related to our work~\citep{lefevre2010cross} use a machine translation based approach to solve the data sparsity in the target language. 
for spoken language understanding systems.

While the above work has been leveraged for multiple tasks, there is no single comparison of the efficacy of these various strategies for contextual understanding and specifically 
reference resolution tasks. In this paper, we provide detailed experiments contrasting and comparing these different approaches for cross-lingual transfer and clearly demonstrate the effectiveness of this approach in both low-resource and large-resource settings.

\section{Conclusion}
\label{sec:conclusion}
In this work, we presented a cross-lingual extension of the {\it context carryover task} for contextual interpretation of slots in a multi-domain large-scale  dialogue system. We explored three different approaches for cross-lingual transfer \textemdash~\ multilingual embeddings, translation projection and delexicalization. Our empirical results on en\_US and de\_DE demonstrate that translation is an effective way to bootstrap systems in a new language. For the context carryover task, specifically, when combining translation with delexicalization, we only need 25\% of the translated data to get within 4\% F1 score of a system trained on true de\_DE annotated collections. We also showed that multilingual embeddings give a small boost in the low data regime, but are not really useful when there is sufficient data, even noisy translated data, in the target domain. 

In the future, we plan to improve our model by leveraging sub-word information which is crucial for robustness to rich morphology in some languages. We also plan on analyzing asian languages like Japanese and Chinese, to see if the above approach generalizes to languages with very different character distributions. Finally, our goal is to reduce reliance on an existence of a translation system, and exploring multi-task objectives where we can learn both language specific and task specific parameters within the same network architecture.

\bibliography{cce2018}
\bibliographystyle{acl_natbib}

\end{document}